\documentclass[11pt,a4paper]{article}
\usepackage{hyperref}
\usepackage{acl2019}
\usepackage{times}
\usepackage{latexsym}
\usepackage{subfiles}
\usepackage{multirow}
\usepackage{graphicx}
\usepackage{algorithm}
\usepackage{url}
\usepackage[noend]{algpseudocode}
\usepackage{arydshln}
\usepackage{booktabs}
\usepackage{color}
\usepackage{colortbl}
\usepackage{amsmath} 
\usepackage{amssymb}
\usepackage{color}
\usepackage[symbol]{footmisc}

\aclfinalcopy % Uncomment this line for the final submission
\title{Enhancing Answer Boundary Detection for Multilingual Machine Reading Comprehension}

\author{
  {Fei Yuan}$^1$\footnotemark[1] \quad {Linjun Shou}$^2$\footnotemark[2] \quad {Xuanyu Bai}$^2$ \quad {Ming Gong}$^2$ \quad {Yaobo Liang}$^3$ \\
  \textbf{Nan Duan}\textsuperscript{3} \quad \textbf{Yan Fu}\textsuperscript{1} \quad \textbf{Daxin Jiang}\textsuperscript{2}\\
  $^1${University of Electronic Science and Technology of China} \\
  $^2${STCA NLP Group, Microsoft, Beijing, China} \\
  $^3${Microsoft Research Asia, Beijing, China} \\
  \texttt{feiyuan@std.uestc.edu.cn}\\
  \texttt{\{lisho,xub,migon,yalia,nanduan,djiang\}@microsoft.com} \\
  \texttt{fuyan@uestc.edu.cn} \\
}

\date{}

\begin{document}
\maketitle

\footnotetext[1]{Work is done during internship at STCA NLP Group, Microsoft.} %对应脚注[1]
\footnotetext[2]{Correspondence author.} 

%%%%%% abstract
%%%%%%%%% ABSTRACT
\begin{abstract}

Multilingual pre-trained models could leverage the training data from a rich source language (such as English) to improve the performance on low resource languages. However, the transfer effectiveness on the multilingual Machine Reading Comprehension (MRC) task is substantially poorer than that for sentence classification tasks, mainly due to the requirement of MRC to detect the word level answer boundary. In this paper, we propose two auxiliary tasks to introduce additional phrase boundary supervision in the fine-tuning stage: (1) a mixed MRC task, which translates the question or passage to other languages and builds cross-lingual question-passage pairs; and (2) a language-agnostic knowledge masking task by leveraging knowledge phrases mined from the Web. Extensive experiments on two cross-lingual MRC datasets show the effectiveness of our proposed approach. 
\end{abstract}

\section{Introduction}
Machine Reading Comprehension (MRC) plays a critical role in the assessment of how well a machine could understand natural language. Among various types of MRC tasks, the span extractive reading comprehension task (like SQuAD~\cite{rajpurkar2016squad}) has been become very popular. Promising achievements have been made with neural network based approaches ~\cite{Seo2017Bidirectional,wang2016multi,xiong2018dynamic,yu2018qanet, hu2017reinforced},
especially those built on pre-trained language models such as BERT ~\cite{devlin2018bert}, due to the availability of large-scale annotated corpora ~\cite{hermann2015teaching,rajpurkar2016squad, joshi2017triviaqa}.
However, these large-scale annotated corpora are mostly exclusive to English, while research about MRC on languages other than English (i.e. multilingual MRC) has been limited due to the absence of sufficient training data.

To alleviate the scarcity of training data for multilingual MRC, the translation based data augmentation approaches were firstly proposed. For example, (question $q$, passage $p$, answer $a$) in English SQuAD can be translated into ($q'$, $p'$, $a'$) in other languages~\cite{asai2018multilingual} to enrich the non-English MRC training data. However, these approaches are limited by the quality of the translators, especially for those low resource languages. 

\begin{table}[t!]
\footnotesize	
\centering
\begin{tabular}{c|cc}
\toprule
\multirow{2}{*}{\textbf{Language}} & MRC & NLI \\
& EM (Gap to English)  & ACC (Gap to English) \\
\midrule
en & 62.4             &  85.0 \\
\cdashline{1-3}[0.8pt/2pt]
es & 49.8 (-12.6)     &  78.9 ( -6.1) \\
de & 47.6 (-14.8)     &  77.8 ( -7.2)\\
ar & 36.3 (-26.1)     &  73.1 (-11.9) \\
hi & 27.3 (-35.1)     &  69.6 (-15.4) \\
vi & 41.8 (-20.6)     &  76.1 ( -8.9)\\
zh & 39.6 (-22.8)     &  76.5 ( -8.5)\\
\bottomrule
\end{tabular}
\caption{The gap between target languages and English on Machine Reading Comprehension (MRC)~\cite{lewis2019mlqa} is significantly larger than sentence level classification task like Natural Language Inference (NLI) ~\cite{conneau2018xnli}. In this experiment, we fine-tune XLM~\cite{conneau2019cross} on English and directly test on other languages.} \label{tab:performance-gap}
% \vspace{-0.4cm}
\end{table}

%%\multirow{2}{*}{\textbf{Language}} & \multicolumn{2}{c}{\textbf{Performance}} \\

Most recently, approaches based on multilingual/cross-lingual pre-trained models~\cite{devlin2018bert,lample2019cross,huang2019unicoder,yang2019xlnet} have proved very effective on several cross-lingual NLU tasks. These approaches learn language-agnostic features and align language representations in vector space during multilingual pre-training process~\cite{wang2019cross, castellucci2019multi,keung2019adversarial, jing2019bipar, cui2019cross}. On top of these cross-lingual pre-trained models, zero-shot learning with English data only, or few-shot learning with an additional small set of non-English data derived from either translation or human annotation, can be conducted. Although these methods achieved significant improvement in sentence level multilingual tasks (like XNLI task~\cite{conneau2018xnli}, the effectiveness on phrase level multilingual tasks is still limited. As shown in  Table \ref{tab:performance-gap}, MRC has bigger gap compared with sentence level classification tasks, in terms of the gap between non-English languages and English. To be specific, the EM metrics for non-English languages have 20+ points gap with the counterpart of English on average.

\begin{table}[t!]
    \centering 
    \footnotesize
    \begin{tabular}{|p{7.5cm}|}
        \hline
        \textbf{[Question]:} who were the kings of the southern kingdom \\
        \textbf{[Passage]:}  In the southern kingdom there was only one dynasty, that of king David, except usurper Athaliah from the northern kingdom, who by marriage, […]\\ 
        \textbf{[Answer - ground truth]:} king David \\
        \textbf{[Answer - model predication:]} David, except usurper Athaliah \\
    
        \hline
        \hline
        \textbf{[Question]:} What is the suggested initial does dosage of chlordiazepoxide \\
        \textbf{[Passage]:} If the drug is administered orally, the suggested initial dose is 50 to 100 mg, to be followed by repeated doses as needed until agitation is controlled – up to 300 mg per day. […]\\ 
        \textbf{[Answer - ground truth]:} 50 to 100 mg \\
        \textbf{[Answer - model predication:]} 100 mg\\    
        \hline
    \end{tabular}
    \caption{Bad answer boundary detection cases of multilingual MRC model.}
    \label{tab:bad-case}
    % \vspace{-0.4cm}
\end{table}

For extractive MRC, the EM metric is very critical since it indicates the answer boundary detection capability, i.e. the accuracy for extractive answer spans. In Table \ref{tab:bad-case}, there are two multilingual MRC cases with wrong boundary detection. In real scenarios, these bad extractive answers will bring negative impact to user experience. One interesting finding after case study is that the multilingual MRC model could roughly locate the correct span but still fail to predict the precise boundary (e.g. missing or adding some words in the spans as the cases in Table \ref{tab:bad-case}). For example, an error analysis of XLM on MLQA~\cite{lewis2019mlqa} showed about 49\% errors come from answers that partially overlap with golden span. Another finding is that a large amount ($\sim{70\%}$ according to MLQA) of the extractive spans are language-specific phrases (kind of broad knowledge, such as entities or N-grams noun phrases). We call such phrases \emph{knowledge phrase} in the rest of paper, and will leverage them as prior knowledge in our model.

Motivated by the above observations, we propose two auxiliary tasks to enhance boundary detection for multilingual MRC, especially for low-resource languages. First, we design a cross-lingual MRC task with mixed-languages $\langle$question, passage$\rangle$ pairs to better align the language representation. We then propose a knowledge phrase masking task as well as a language-agnostic method to generate per-language knowledge phrases from the Web. Extensive experiments on two multilingual MRC datasets show that our proposed tasks could substantially boost the model performance on answer span boundary detection. The main contributions of our paper can be summarized as follows.
\begin{itemize}
\item We design two novel auxiliary tasks in multi-task fine-tuning to help improve the accuracy of answer span boundary detection for multilingual MRC model.
\item We propose a language-agnostic method to mine language-specific knowledge phrase from search engines. This method is light-weight and easy to scale to any language.
\item We conduct extensive experiments to prove the effectiveness of our proposed approach. In addition to an open benchmark dataset, we also create a new multilingual MRC dataset from real-scenario together with fine-grained answer type labels the in-depth impact analysis. 
\end{itemize}

\begin{figure*}[t!]
    \centering
    \includegraphics[trim={1.6cm 4.5cm 3.2cm 3cm},clip,scale=0.55]{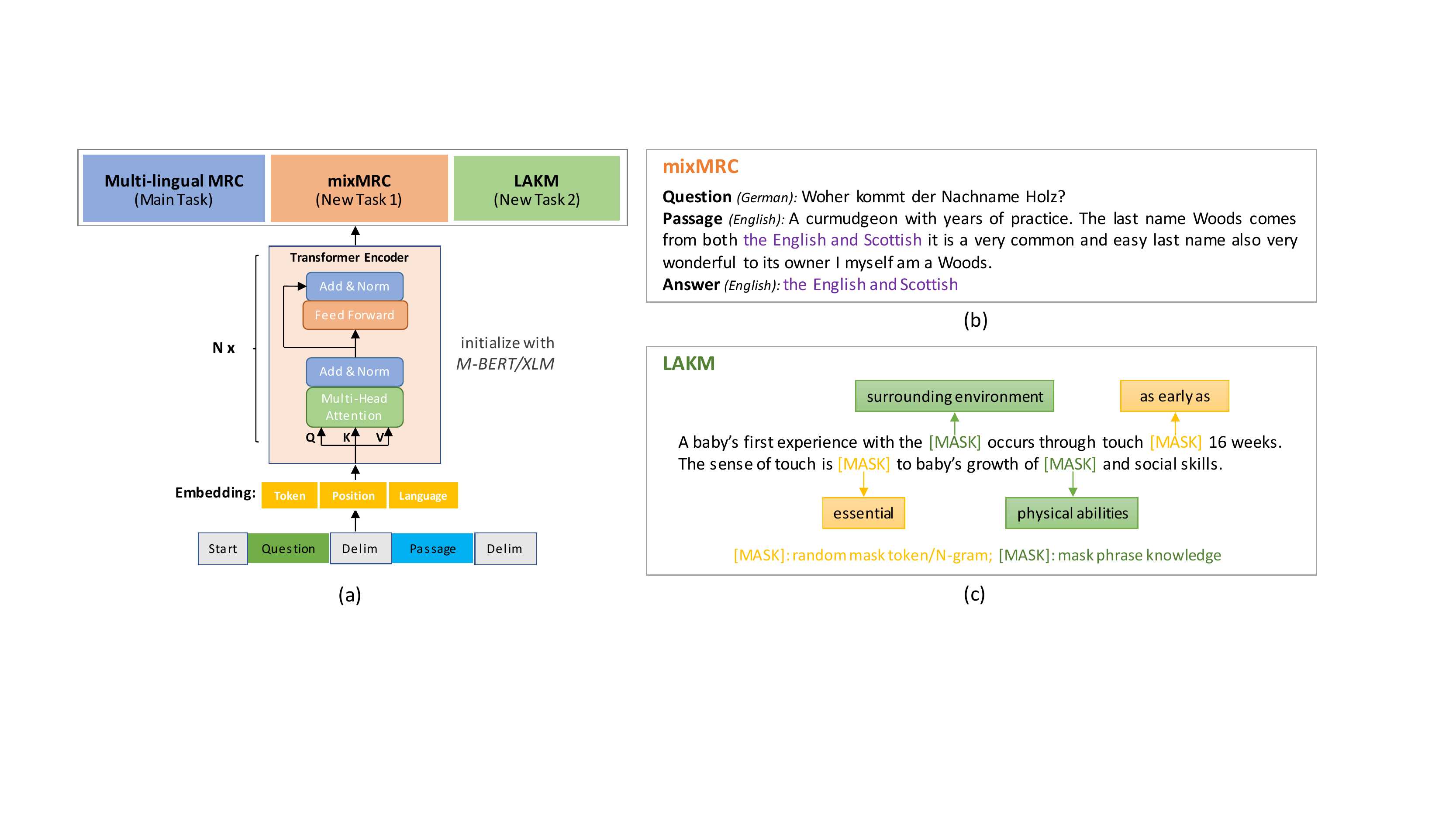}
    \caption{Overview of enhancing answer boundary detection work for multilingual machine reading comprehension. Our approach consists of three tasks: (a) Main task: multilingual MRC model requires to read text material and answer the question based on given context; (b) mixMRC task: cross-lingual MRC task with mix-language $\langle$question, passage$\rangle$ pairs; (c) LAKM task: A language-agnostic knowledge masking task by leveraging language-specific knowledge mined from web.}
    \label{fig:overview}
    % \vspace{-0.4cm}
\end{figure*}

\section{Related Work}

\subsection{Multilingual Natural Language Understanding (NLU)} 
A straightforward approach is leveraging translation to translate training data in rich resource language to low resource language. \citeauthor{asai2018multilingual} (\citeyear{asai2018multilingual}) proposed to use run-time machine translation for multilingual extractive reading comprehension. \citeauthor{cui2019cross} (\citeyear{cui2019cross}) developed several back-translation methods for cross-lingual MRC. \citeauthor{singh2019xlda} (\citeyear{singh2019xlda}) introduced a translation-based data augmentation mechanism for question answering. However, these methods highly depend on the availability and quality of translation systems.

Another approach to Multilingual NLU extracts language-independent features to address multilingual NLU tasks. Some works~\cite{keung2019adversarial,jia2017adversarial,chen2019multi} apply adversarial technology to learn language-invariant features and achieve significant performance gains. More recently, there has been an increasing trend to design cross-lingual pre-trained models, such as multilingual BERT~\cite{devlin2018bert}, XLM~\cite{lample2019cross}, and Unicoder~\cite{huang2019unicoder}, which showed promising results due to the capability of cross-lingual representations in a shared contextual space ~\cite{pires2019multilingual}. In this paper, we propose two novel sub-tasks in fine-tuning cross-lingual models for MRC.

\subsection{Knowledge based MRC} 

Prior works~\cite{yang-mitchell-2017-leveraging, mihaylov2018knowledgeable, weissenborn2017dynamic, sun2018knowledge} mostly focus on leveraging structured knowledge from knowledge bases (KBs) to enhance MRC models following a retrieve-then-encode paradigm, i.e., relevant knowledge from KB are retrieved first and sequence modeling methods are used to capture complex knowledge features. However, such a paradigm often suffers from the sparseness of knowledge graphs. 

Recently, some works fuse knowledge into pre-trained models to get knowledge enhanced language representation. \citet{zhang2019ernie} uses both large-scale textual corpora and knowledge graphs to train an enhanced language representation. \citet{sun2019ernie} construct unsupervised pre-trained tasks with large scale data and prior knowledge  to help the model efficiently learn the lexical, syntactic and semantic representations, which significantly outperforms BERT on MRC.   

Most previous works on knowledge-based MRC are limited to English only. Meanwhile the requirement of acquiring large-scale prior knowledge (such as entity linking, NER models) may be challenging to meet for non-English languages. In this work, we propose a light-weight language-agnostic knowledge phrase mining approach and design a knowledge phrase masking task to boost the model performance for {\em multilingual} MRC. 

\begin{figure*}[t!]
    \centering
    \includegraphics[trim={1.2cm 10cm 2cm 6cm},clip,scale=0.65]{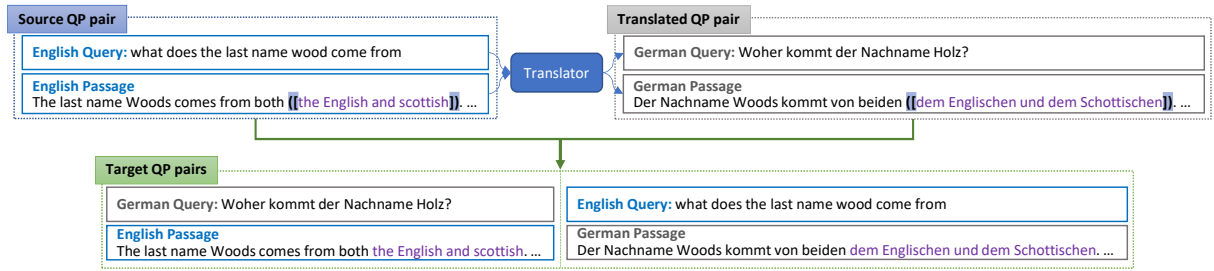}
    \caption{MixMRC data generation process. Given source (English) QP pair, we translate QP pair from English into non-English. Then the target mix-language pair can be divided into two forms: translated question-source passage and source question and translated passage pair.}
    \label{fig:mix_translation_data_generation}
\end{figure*}

\section{Approach}
In this section,  we first introduce the overall training procedure, and then introduce two new tasks, namely, Mixed  Machine  Reading  Comprehension (mixMRC) and Language-agnostic Knowledge Phrase Masking (LAKM), respectively.

The overview of our training procedure is shown at Figure~\ref{fig:overview}. Our approach is built on top of popular multilingual pre-trained models (such as multilingual BERT and XLM). We concatenate passage, question (optional) together with special tokens {\tt [Start]} and {\tt [Delim]} as the input sequence of our model, and transform word embedding into contextually-encoded token representations using transformer. Finally, this contextual representation is used for all three tasks introduced as following.

The first task, also our main task, is multilingual MRC, which aims to extract answers spans from the context passage according to the question. In this task, each language has its own data. However, only English has human labeled training data, and the other languages use machine translated training data from English. During training, the MRC training data in all languages will be used together for fine-tuning.

In the following, we introduce our new proposed tasks which will jointly train with our main task to boost multilingual MRC performance.
% Additionally, we propose two auxiliary tasks to enhance the answer boundary extraction for multilingual MRC. This section will describe the details, including its training data generation process and model structure about these auxiliary tasks.

\subsection{Mixed Machine Reading Comprehension (mixMRC)}

We propose a task, named mixMRC, to detect answer boundaries even when $\langle$question, passage$\rangle$ are in different languages, which is shown in Figure~\ref{fig:overview} (b). It is mainly motivated by the strategy of data augmentation~\cite{singh2019xlda}. In detail, we utilize the mixMRC to derive more accurate answer span boundaries according to the constructed $\langle$question, passage$\rangle$ pairs.

The way to obtain $\langle$question, passage$\rangle$ pairs consists of two steps: 1) translate training data from English into non-English; 2) construct mix-language training data for mix-MRC task. We show the entire data generation process in the Figure~\ref{fig:mix_translation_data_generation}.

\paragraph{Step 1: Data Translation}
When using machine translation system to translate paragraphs and questions from English into non-English, the key challenge is how to address the answer span in translation.

% we find the answer span tends to be lost in the process of translation.

% To minimize the skip ratio in translation, 
To solve this problem, we enclose the answer text of source passage in special token pair "([" and "])", similar to ~\cite{lee_semi-supervised_2018}. After translation, we discard training the instances where the translation model does not map the answer into a span well. Some skip data can still be recalled by finding the translated answer in the translated passage. The statistics of translated data are shown in Table~\ref{tab:translation-performance}. 

Formally, given a monolingual dataset $D = \{(q_i, p_i, a_i)\}$ where  $q_i$, $p_i$ and $a_i$ mean the query, passage and answer of language $i$ respectively. We apply a public translator and create a translated dataset $D' = \{(\tilde{q_j}, \tilde{p_j}, \tilde{a_j})\}$, where $\tilde{q_j}$ is the translation of $q_i$, and $\tilde{a_j}$ is the answer span boundary in $\tilde{p_j}$. 
%  and $j \neq k$
\begin{table}[h]
    \centering
    \small
    \begin{tabular}{c|cc|cc}
        \toprule
          &  \multicolumn{2}{c|}{\textbf{MTQA}} & \multicolumn{2}{c}{\textbf{MLQA}} \\
         & \# instance & skip ratio & \# instance & skip ratio \\
         \midrule
         en & 56616 & - & 87599 & - \\
         fr & 52502 & 0.0727 & - & - \\
         de & 51326 & 0.0934 &  80284 & 0.0835 \\
         es & -& - & 87134 & 0.0053 \\
        \bottomrule
    \end{tabular}
    \caption{\label{tab:translation-performance} The statistics of translated data. The skip ratio is the percentage of those cases which are discarded.}
    % \vspace{-0.4cm}
\end{table}

\paragraph{Step 2: Mix Language}
After translation, we create a mixed-language dataset  $D'' = \{(\tilde{q_k}, \tilde{p_l}, \tilde{a_l})\}$ where $l \neq k$. This could encourage MRC model to distinguish the phrases boundary by answer span selection and also keep the alignment of the underlying representations between two languages. In this task, we use the same fine-tuning framework as in monolingual MRC task.

% The process of modeling answer-ability of $(\tilde{q_j}, \tilde{p_k}, \tilde{a_k})$ from mix-language dataset $D'$ is similar to pre-trained model fine-tuning on SQuAD.

\begin{figure*}[t!]
    \centering
    \includegraphics[trim={2cm 6cm 2.5cm 5.3cm},clip,scale=0.57]{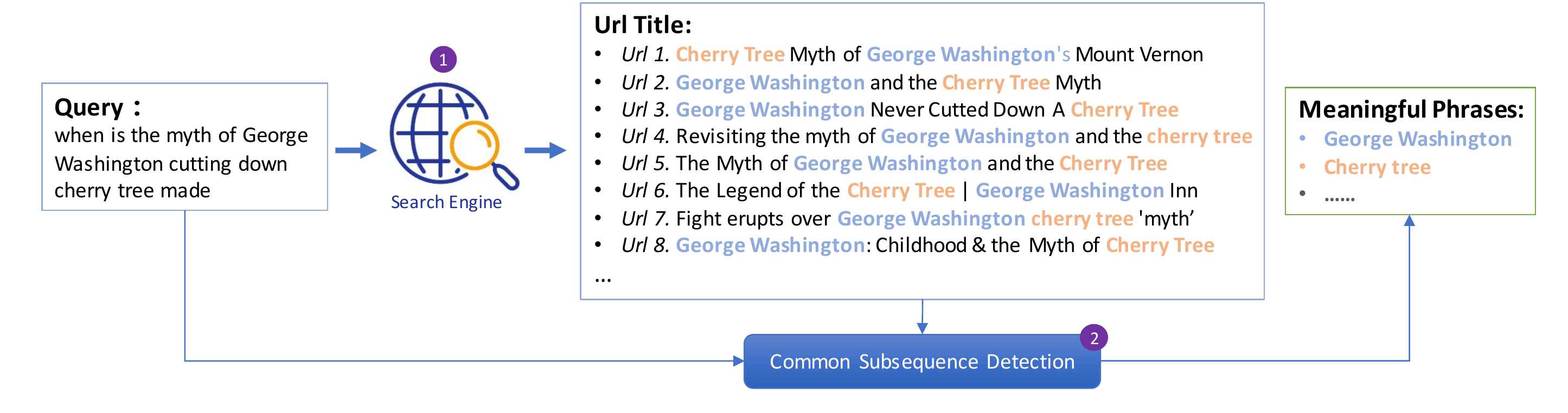}
    \caption{The process to generate knowledge data.}
    \label{fig:process for knowledge}
    % \vspace{-0.4cm}
\end{figure*}

\subsection{Language-agnostic Knowledge Phrase Masking (LAKM)}
In this section, we first introduce the approach for mining knowledge phrases from the Web. We then introduce the masking task created with these knowledge phrases.

% shows the language-agnostic knowledge masking (LAKM) task. This task uses  These knowledge phrases can be further leveraged for model training to enhance answer boundary detection, because they contain language-specific information.

\paragraph{Data Generation}
In the following, we will describe our data generation method to collect large-scale phrase knowledge for different languages. The source data comes from a search engine, consisting of queries and the top N relevant documents. Let us take a running example of query {\tt \{when is the myth of George Washington cutting down cherry tree made\}}. As shown in Figure~\ref{fig:process for knowledge}, our mining pipeline consists of two main steps:

\begin{enumerate}
    \item  Phrase Candidates Generation: This step targets at high recall. We enumerate all the n-grams (n=2,3,4) of the given query as phrase candidates, such as {\tt  when is, the myth, George Washington, cherry tree, is the myth, etc}.  We further filter the candidates with a stop word list.  A manual analysis (by asking humans to identify all meaningful n-gram phrases in the given queries) shows that recall reaches $\sim{83\%}$.
    \item  Phrase Filtering: This step targets at high precision by removing useless phrases. For each candidate, we count its frequency in the titles of relevant documents. We only keep those frequent candidates. For example, phrases {\tt George Washington, cherry tree} appear in every title. We name them as \emph{knowledge phrases}. Our empirical study suggests a frequency of 0.7 results in a good balance between precision and recall, and we use this threshold in our approach.
\end{enumerate}

Following this approach, large amount of meaningful phrases can be mined independent of languages. After this, we further extract the passages which contain the mined knowledge phrases from the documents (following similar passage creation approach proposed by \citet{rajpurkar2016squad}), which is the input of the LAKM. For the purpose of fair comparisons, the number of passages in different languages is equal, and the total amount of training data in LAKM is the same as that of mixMRC. The statistics of the knowledge phrases are given in Table ~\ref{tab:knowledge-info}. 

\begin{table}[h]
\small
    \centering
    \begin{tabular}{p{3.5cm}p{0.5cm}p{0.5cm}p{0.5cm}p{0.5cm}}
    \toprule
         & \textbf{en} &\textbf{ fr} & \textbf{de} &\textbf{ es} \\
         \midrule
         \# passages & 99.7k & 91.2k & 93.8k & 78.8k \\
         \# knowledge phrases & 229k & 102k & 102k & 101k \\
         Avg. knowledge words & 2.14 & 2.36 & 2.18 & 2.19\\
         Avg. knowledge / passage & 2.29 & 1.11 & 1.09 & 1.28 \\
    \bottomrule
    \end{tabular}
    \caption{Statistics of the knowledge data we used.}
    \label{tab:knowledge-info}
    % \vspace{-0.4cm}
\end{table}

\paragraph{Model Structure}
Given a $\langle$passage, knowledge phrases$\rangle$ pair, denoted as $(X, Y)$, we formalize that $X = (x_1, x_2, …, x_m)$ is a passage with $m$ tokens, $Y = (y_1, y_2, …, y_n)$ is a set of language-specific knowledge phrases generated as before, where $y_i = (x_j, x_{j+1}, …, x_{j+(l-1)}) (1 \leq j \leq m)$, $l$ is the number of tokens in  $y_i ( 1 \leq i \leq n)$. The representations $h_\theta$ can be easily obtained from transformer. To inject language-specific knowledge into multilingual MRC model, we use masked language model as the fine-tuning objective. This task-specific loss has an additional summation over the length of sequence:

\begin{equation}
    p_t =  \textrm{Softmax} (Wh_\theta(x)_t + b)
\end{equation}
\begin{equation}
     L_{LAKM} = \sum_{k=1}^m{-y_{kt}^T\textrm{log}p_t}
\end{equation}
where $p_t$ is the prediction value of $t^{th}$ word, $m$ is the number of tokens in the input passage, $y_{kt}$ is the target word, $W, b$ are the output projections for the task-specific loss $L_{LAKM}$, and $h_\theta(x)_t$ refers to the pre-trained embedding of the $t^{th}$ word.

\section{Experiments}
In this section, we firstly describe the dataset and evaluation in Section \ref{dataset}; then introduce the baseline models in Section \ref{baseline} and experiment setting in Section \ref{exp_setting}; thirdly the experimental results are shown in Section \ref{exp_results}.

% on two multilingual MRC datasets: Multi-Type Question Answering (MTQA) and MLQA~\cite{lewis2019mlqa}. 

\begin{figure*}
    \centering
    \includegraphics[trim={0cm 0cm 0.3cm 0cm},clip,scale=0.33]{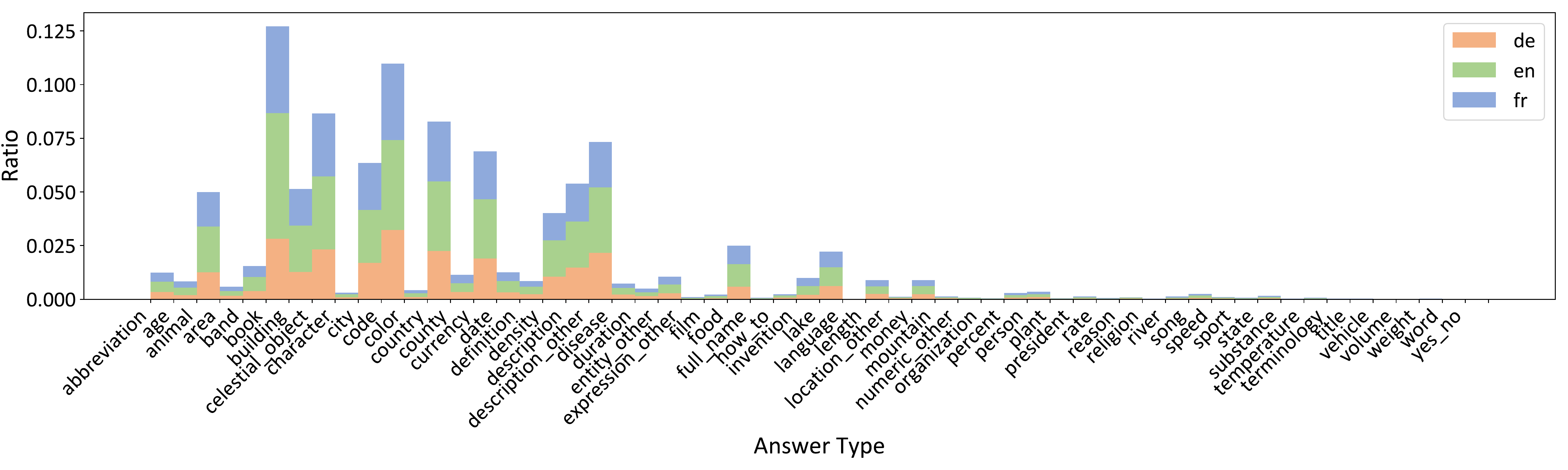}
    \caption{\label{fig:answer_type_distribution} Answer type distribution in MTQA.}
    % \vspace{-0.2cm}
\end{figure*} 

\subsection{Dataset and Evaluation}
\label{dataset}
    
\subsubsection{Dataset}
To verify the effectiveness of our approach, we conduct experiments on two multilingual datasets: one open benchmark called MLQA~\cite{lewis2019mlqa}; the other newly constructed multilingual QA dataset with multiple fine-grained answer types (MTQA).

    \paragraph{MLQA.} 
    A multilingual question answering benchmark~\cite{lewis2019mlqa}. MLQA contains QA instances in 7 languages. Due to resource limitation, we evaluate our models on three languages (\emph{English, German, Spanish}) of the dataset. 
  
    \paragraph{MTQA.}  
    To further evaluate our approach on real-scenario as well as conduct in-depth analysis of the impact on different answer types (in Section \ref{extensive}), we construct a new QnA dataset with fine-grained answer types. The construction process is described as following: 
    \begin{enumerate}
        \item $\langle$question, passage$\rangle$ pairs come from the question answering system of one commercial search engine. Specifically, questions are real user searched queries on one commercial search engine, which are more diverse, covering various answer types. For each question, a QA system is leveraged to rank the best passage from the top 10 URLs returned by search engine. For each question, only the best passage is selected.
        
        \item To annotate the answer span in each passage, we leverage crowd sourcing annotators for the labeling. Annotators are asked to first select the best shortest span\footnote{Only single span is considered.} in the passage which can answer the question and also assign an answer type according to the query and the answer span. Each case are labeled by three annotators and those instances which are labeled with consensus (no less than two annotators agree on the result) are finally selected.  An English example is given in Table \ref{tab:case-MTQA}.
    \end{enumerate}

Detailed statistics of MTQA dataset are given in Table~\ref{tab:stat-dataset} as well as the distribution of answer types in our dataset shown in Figure~\ref{fig:answer_type_distribution}.
    
\begin{table}[h]
    \centering 
    \footnotesize
    \begin{tabular}{p{7.0cm}}
        \hline
        \textbf{[Question]:} how many players in rugby-league team on field \\
        \textbf{[Passage]:}  A rugby league team consists of thirteen players on the field, with four substitutes on the bench, […]\\ 
        \textbf{[subtype]:} numeric \\
        \textbf{[Answers:]}"start":41,"end":49,"text":"thirteen" \\
        \hline
    \end{tabular}
    \caption{An English example of MTQA.}
    \label{tab:case-MTQA}
    % \vspace{-0.4cm}
\end{table}

\begin{table}[h]
	\centering
	\small
	\begin{tabular}{p{3.3cm}|ccc}
	    \toprule
		 & \textbf{en} & \textbf{fr} & \textbf{de} \\
		 \midrule
		\# of dev instances &  6156 & 4900 & 3975\\
		\# of test instances & 3017 & 2413 & 1893 \\
		\# of dev answer type &  58 & 57 & 55\\
		\# of test answer type & 54 & 51 & 53\\
		\bottomrule
	\end{tabular}
	\caption{\label{tab:stat-dataset} Statistics of the dataset MTQA.}
% 	\vspace{-0.4cm}
\end{table}
  
% All models they used were performed using the Pytext NLP toolkit~\cite{aly2018pytext}. However all models we used were performed using hugging face toolkit ~\cite{Wolf2019HuggingFacesTS}. Other than this one change, we use Google Neural Machine Translation (GNMT) system.

\subsubsection{Experimental Evaluation}
We use the same evaluation metrics in the SQuAD dataset~\cite{rajpurkar2016squad}, i.e., \emph{F1} and \emph{Exact Match}, to evaluate the model performance.
Exact Match Score measures the percentage of predictions that exactly match any one of the ground truths. F1 score is used to measure the answer overlap between predictions and ground truth. We treat the predictions and ground truth as bags of words, and compute their F1 score. For a given question, we select the maximum value of F1 over all of the ground truths, and then we average over all of the questions.

% \begin{itemize}
%     \item \textbf{Exact Match Score:}
%     This metric measures the percentage of predictions that exactly match any one of the ground truths. 
%     \item \textbf{F1 Score:}
% 	This metric is used to measure the answer overlap between predictions and ground truth. We treat the predictions and ground truth as bags of words, and compute their F1 score. For a given question, we select the maximum value of F1 over all of the ground truths, and then we average over all of the questions.
% \end{itemize}

\subsection{Baseline Models}
\label{baseline}
We use the following two multilingual pre-trained models to conduct experiments:
\begin{itemize}
    \item \textbf{M-BERT:} Multilingual version of BERT released by \cite{devlin2018bert} which is pre-trained with monolingual corpora in 104 languages. This model proves to be very effective at zero-shot multilingual transferring between different languages \cite{pires2019multilingual}.
    \item \textbf{XLM:} A cross-lingual language model (15 languages)~\cite{lample2019cross} pre-trained with both monolingual data and cross-lingual data as well as cross-lingual tasks to enhance the transferring capacity among different languages. 
\end{itemize}

For baseline, we directly fine-tune the pre-trained models using MRC training data only. 

% We present experimental results on multilingual BERT (cased, 104 languages) \cite{devlin2018bert} and XLM (MLM + TLM, 15 languages) \cite{lample2019cross}. 

% In Table~\ref{tab:translation-performance}, en means English dataset, which has human labeled training data, and the other languages use machine translated training data from English. After translation,  MTQA consists of 56616 extractive QA instances in English (en), 52502 instances in French (fr), and 51326 instances in German (de). MLQA is made up of 87599 instances in English (en), 80284 instances in German (de) and 87134 instances in Spanish (es).

\begin{table*}[t!]
\small
\centering
\begin{tabular}{p{1.0cm}p{2.6cm}|ccc|ccc}
\toprule
%% methods | dataset_1 | dataset_2
\multirow{2}{*}{\bf{Model}} & \multirow{2}{*}{\bf{Methods}} & \multicolumn{3}{c|}{\textbf{MLQA} (EM / F1)} & \multicolumn{3}{c}{\textbf{MTQA} (EM / F1)} \\
 &  &  en & es & de & en & fr & de \\
\midrule
 \multirow{5}{*}{M-BERT}
& \citet{lewis2019mlqa}  & 65.2         / 77.7 &          37.4 / 53.9 &         47.5 / 62.0  &    -                 &                    - &   -         \\
& Baseline   & 65.4         / 79.0 &          50.4 / 68.5 &         46.2 / 60.6  &          67.0 / 86.9 &          52.9 / 78.2 & 59.8 / 81.4 \\
% \hline{2-8}
& LAKM      &\textbf{66.9} / 80.1 &          51.5 / 69.5 &         49.9 / 64.4  & \textbf{68.8} / 87.6 &          56.8 / 78.8 & 62.4 / 81.9 \\
& mixMRC      & 65.4         / 79.4 &          50.5 / 69.1 &         49.1 / 64.0  &          67.9 / 86.8 &          56.4 / 77.8 & 62.4 / 81.0 \\
& mixMRC + LAKM & 64.7         / 79.2 & \textbf{52.1} / 70.4 & \textbf{50.9} / 65.6 &          68.6 / 87.0 & \textbf{57.5} / 78.5 & \textbf{62.9} / 81.3 \\ 
\midrule
\multirow{5}{*}{XLM }
& \citet{lewis2019mlqa}  &         62.4 / 74.9 &          47.8 / 65.2 &          46.7 / 61.4 &          -           & -                    &          -           \\
& Baseline   &         64.1 / 77.6 &          50.4 / 68.4 &          47.4 / 62.0 &          67.1 / 86.8 &          51.5 / 75.8 &          61.6 / 81.3 \\
% \hline{2-8}
& LAKM          &\textbf{64.6} / 79.0 &          52.2 / 70.2 &          50.6 / 65.4 & \textbf{68.3} / 87.3 &          52.5 / 75.9 &          61.9 / 81.2 \\ 
& mixMRC      &         63.8 / 78.0 &          52.1 / 69.9 &          49.8 / 64.8 &          66.5 / 85.9 &          52.9 / 75.0 &          62.1 / 80.5 \\
& mixMRC + LAKM &         64.4 / 79.1 & \textbf{52.2} / 70.3 & \textbf{51.2} / 66.0 &          68.2 / 86.8 & \textbf{53.6} / 75.9 & \textbf{62.5} / 80.9 \\
\bottomrule 
\end{tabular} 
\caption{\label{tab:experimental-result} Experimental results on MLQA and MTQA dataset under translation condition (\%).} 
% \vspace{-0.2cm}
\end{table*}
	
\subsection{Experimental Setting}
\label{exp_setting}
    We use Adam optimizer with $\beta_1 =0.9$ , $\beta_2=0.999$. The learning rate is set as 3e-5 for the mixMRC, LAKM and multilingual MRC tasks. The pre-trained model is configured with its default setting. Each of the tasks is trained until the metric of MRC task converges. 
    \paragraph{mixMRC.}
    We jointly train mixMRC and multilingual MRC tasks using multi-task training at the batch level to extract the answer boundary in the given context. For both tasks, the max sequence length is 384.
    
    \paragraph{LAKM.}
    LAKM and multilingual MRC tasks are jointly trained using multi-task training. In terms of input, we randomly mask 15\% of all WordPiece tokens in each sequence in a two step approach. Firstly, if the $i-th$ token belongs to a knowledge phrase, we replace the $i$- token with (1) the {\tt [MASK]} token 80\% of the time (2) a random token 10\% of the time (3) the unchanged $i-th$ token 10\% of the time. Secondly, if the proportion of knowledge phrase is less than 15\%, we will further randomly mask other WordPiece tokens to make the total masked ratio to reach 15\%. For LAKM, the max sequence length is set as 256. 

    \paragraph{mixMRC + LAKM.}
    We jointly train mixMRC, LAKM and multilingual MRC tasks, take the gradients with respect to the multilingual MRC loss, mixMRC loss and LAKM loss, and apply the gradient updates sequentially at batch level. During the training, the max sequence length is 384 for multilingual MRC model, 256 for LAKM and 384 for mixMRC.

\subsection{Experiment Results}
\label{exp_results}

The overall experimental results are shown in Table~\ref{tab:experimental-result}. Compared with M-BERT \& XLM baselines, both mixMRC and LAKM have decent improvements in fr, es and de, and on-par performance in en  in terms of both MLQA and MTQA datasets. This demonstrates the effectiveness of our models.

The combination of LAKM and mixMRC tasks gets the best results on both datasets. Take M-BERT and MLQA dataset as an example, mixMRC+LAKM have 1.7\% and 4.7\% EM improvements on es and de languages respectively, compared with baseline.  

In terms of LAKM task, there are decent gains for all languages, including English. However, the gains are bigger on low resource languages compared with English performance. Take XLM and MLQA dataset as an example, LAKM gets 1.8\% and 3.2\% EM improvements on es and de, while the improvement on en is about 0.5\%. The intuition behind en gains is that LAKM brings extra data with knowledge to en as well. 

In terms of mixMRC task, there are slight regression on en compared with decent gains on es, de and fr. Take XLM and MTQA dataset for illustrations, mixMRC has 0.6\% EM regression on en versus 1.4\% and 0.5\% EM gains on fr and de languages. This shows that mixMRC mainly improves the transferring capability from rich resource language to low resource language.   

% there are bigger gains on low resouce languages com In other languages, we also can see a boosting performance in es (AVG. 0.9\%) and de (AVG. 2.1\%). However, the performance in English does not increase obviously at the assistance of mixMRC task.

% Comparing the tasks of LAKM and mixMRC, we can see that for MLQA, the exact match value is improved by  -0.3\%, 1.7\%, 2.4\% using XLM-Based mixMRC task, while we obtain larger improvement with 0.5\%, 1.8\%, 3.2\% demonstrating the effectiveness of the proposed LAKM method. Meanwhile, we can find there is slight regression on English result compared with LAKM. 

\section{Analysis}
In this section, we ablate important components in LAKM to explicitly demonstrate its effectiveness. 

\subsection{Random N-gram Masking vs LAKM}
To study the effectiveness of LAKM, we compare LAKM with \emph{Random N-gram Masking}\footnote{Random N-gram Masking shows gains in English SQuAD.} based on XLM and MTQA dataset. LAKM and Random N-gram Masking refer to fine-tuning XLM with the language-specific knowledge masking strategy and random n-gram masking strategy respectively. As shown in Table ~\ref{tab: model-structure}, without the language-agnostic knowledge masking strategy, the EM metrics drops by 0.2\% - 0.87\%, which proves the necessity of LAKM. 

\begin{table}[htbp]
\small
\centering
\begin{tabular}{p{3.5cm}ccc}
%% methods | dataset_1 | dataset_2
\toprule
\textbf{Setting} (EM) & \textbf{en}  & \textbf{fr} & \textbf{de} \\
\midrule
% Random N-gram Masking & 67.45 & 51.82 & 61.73 \\
% LAKM & 68.32 & 52.46 & 61.93 \\
Random N-gram Masking & 67.5 & 51.8 & 61.7 \\
LAKM & 68.3 & 52.5 & 61.9 \\
\bottomrule
\end{tabular}
\caption{Ablation study on MTQA (\%).} \label{tab: model-structure}
% \vspace{-0.4cm}
\end{table}

\subsection{Zero Shot Fine-tuning w/ vs w/o LAKM} 
To illustrate the effectiveness of the auxiliary tasks, an extreme scenario is considered when only English training data is available and there is no translation data. That means that we are unable to use mixMRC task to driver more accurate answer span boundaries. At this point, we only leverage LAKM to enhance answer boundary detection and compares the performance of M-BERT baseline with our model in  Table~\ref{tab:zero-shot-result}.

From the experimental results, zero shot fine-tuning with LAKM is significantly better than M-BERT baseline. On MTQA, our model gets 2\%, 3.3\%, 3.8\% EM improvements on English, French and German respectively. On MLQA, we get 1.6\%, 1.4\%, 1.2\% EM improvements on English, Spanish and German.

\iffalse
\begin{table}[h]
    \toprule
    \small
    \begin{tabular}[t]{p{0.3cm}|cc|cc}
    \centering
    %% methods | dataset_1 | dataset_2
    & \multicolumn{2}{c|}{\textbf{MLQA} (EM / F1)} & \multicolumn{2}{c}{\textbf{MTQA} (EM / F1)}  \\
    & w/o & w/  & w/o  & w/  \\
    \midrule
    en &  65.2 / 77.7 & \textbf{66.8 / 80.0} &  65.8 / 86.6 & \textbf{67.8 / 87.2} \\
    fr &  - & -  &  41.3 / 70.9 & \textbf{44.6 / 72.1} \\
    de &  44.3 / 57.9 & \textbf{45.5 / 60.5} & 50.7 / 76.2 & \textbf{54.5 / 77.8} \\
    es &  46.6 / 64.3 & \textbf{ 48.0 / 65.9 } & - &  -   \\
    \bottomrule
    \end{tabular} 
\caption{\label{tab:zero-shot-result} Experimental results on MLQA and MTQA datasets (\%). w/o refers to only use English data to train MRC model; w/ means using LAKM assists the English MRC model training.} 
% \vspace{-0.4cm}
\end{table}
\fi
\begin{table}[h]
    
    \small
    \begin{tabular}[t]{l|ccc}
    \toprule
    \centering
    %% methods | dataset_1 | dataset_2
    & \multicolumn{3}{c}{\textbf{MLQA} (EM / F1)} \\
    & en & es & de  \\
    \midrule
    Baseline & 65.2 / 77.7 & 46.6 / 64.3 & 44.3 / 57.9 \\
    LAKM & \textbf{66.8 / 80.0} & \textbf{ 48.0 / 65.9 } & \textbf{45.5 / 60.5} \\
    \midrule
    \multicolumn{4}{c}{} \\
    \midrule
    & \multicolumn{3}{c}{\textbf{MTQA} (EM / F1)} \\
    & en & fr & de  \\
    \midrule
    Baseline & 65.8 / 86.6 & 41.3 / 70.9 & 50.7 / 76.2 \\
    LAKM & \textbf{67.8 / 87.2} & \textbf{44.6 / 72.1} & \textbf{54.5 / 77.8} \\
    \bottomrule
    \end{tabular} 
\caption{\label{tab:zero-shot-result} Zero Shot experimental results on MLQA and MTQA datasets (\%). We only use English MRC training data and don't use translation data.} 
% \vspace{-0.4cm}
\end{table}
    
\subsection{Extensive Analysis on Fine-grained Answer Types}
\label{extensive}
\begin{figure}[h]
    \centering
    \includegraphics[trim={3.2cm 2.3cm 11cm 2cm},clip,scale=0.28]{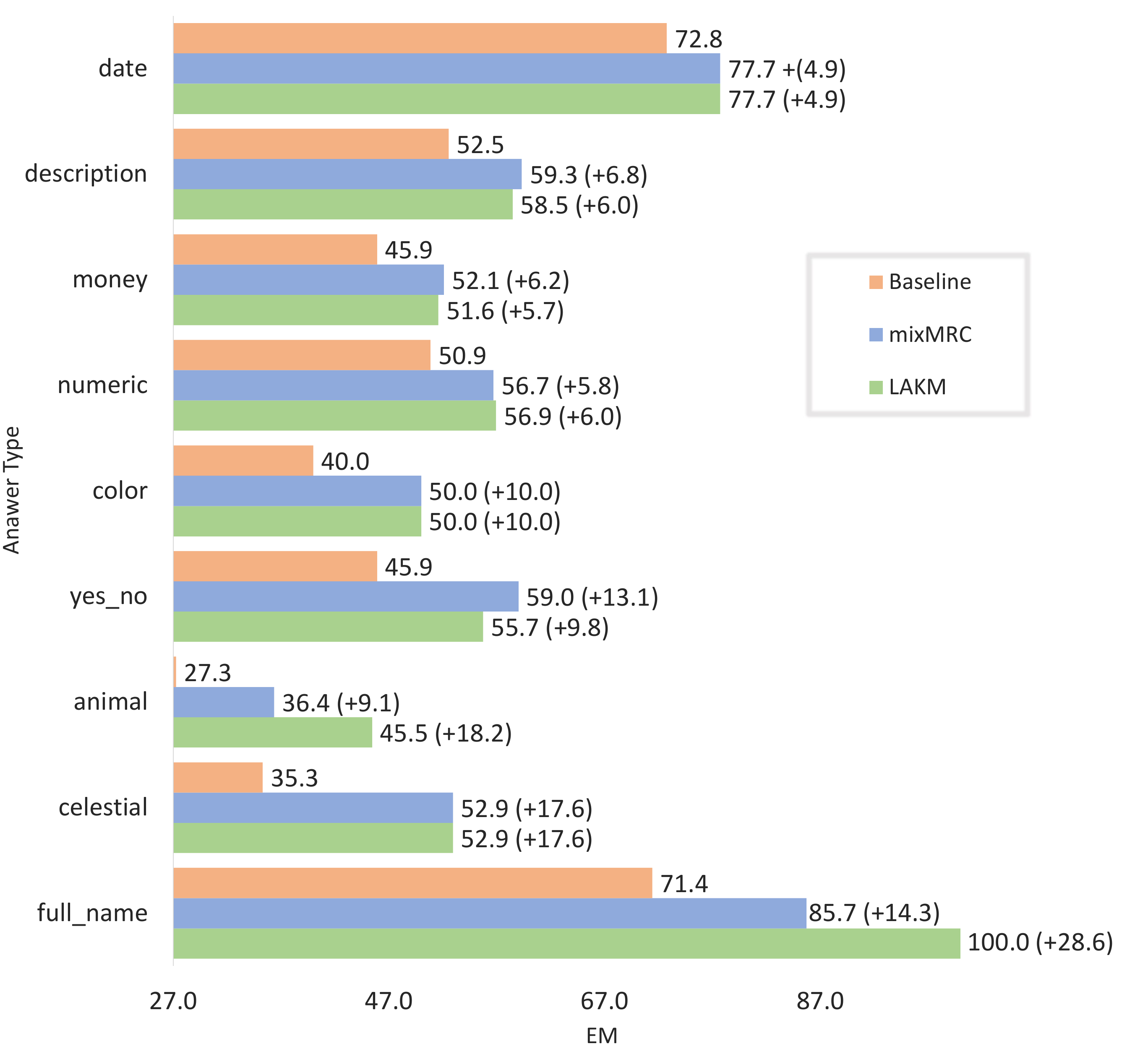}
    \caption{EM results comparison on M-BERT (MTQA French test set) for the different answer types.}
    \label{fig:error-analysis}
    % \vspace{-0cm}
\end{figure}

To have an insight that how the new tasks (LAKM/mixMRC) affect the multilingual MRC task, we further analyze model performance on various answer types, as shown in Figure~\ref{fig:error-analysis}.

The comparison with baseline indicates that in most of the answer types (like {\tt color, description, money}), both LAKM and mixMRC can enhance the answer boundary detection for multilingual MRC task. 

One interesting finding is that in terms of {\tt animal, full\_name}, LAKM outperforms mixMRC by a great margin, which are 9.1\% and 14.3\% respectively. One possible explanation is that the knowledge phrases of LAKM can cover some entity related phrases like animals and names, leading to the significant EM boost.

In terms of those numerical answer types (like {\tt money, numeric, length}), the performance between mixMRC and LAKM are similar. The intuition behind this is that these numerical answers may be easier to transfer between different languages since answers like length are similar across different languages.  

\section{Conclusion}

This paper proposes two auxiliary tasks (mixMRC and LAKM) in the multilingual MRC fine-tuning stage to enhance answer boundary detection especially for low resource languages. Extensive experiments on two multilingual MRC datasets have been conducted to prove the effective of our proposed approach. Meanwhile, we further analyze the model performance on fine-grained answer types, which shows interesting insights.  

%%%%%% Reinference
\bibliography{mrc_paper}
\bibliographystyle{acl_natbib}

\end{document}